# Large language models in medicine: the potentials and pitfalls

Jesutofunmi A. Omiye[1]*, Haiwen Gui[1]*, Shawheen J. Rezaei[1], James Zou[2], Roxana Daneshjou[1,2]
*These authors contributed equally as a co-first author to this manuscript
[1]Department of Dermatology, Stanford University, Stanford, USA
[2]Department of Biomedical Data Science, Stanford University, Stanford, USA

**Abstract:**
Large language models (LLMs) have been applied to tasks in healthcare, ranging from medical exam questions to responding to patient questions. With increasing institutional partnerships between companies producing LLMs and healthcare systems, real world clinical application is coming closer to reality. As these models gain traction, it is essential for healthcare practitioners to understand what LLMs are, their development, their current and potential applications, and the associated pitfalls when utilized in medicine. This review and accompanying tutorial aim to give an overview of these topics to aid healthcare practitioners in understanding the rapidly changing landscape of LLMs as applied to medicine.

## 1. Introduction:

Large Language models (LLMs) have become increasingly mainstream since the launch of OpenAI's (San Francisco, USA) publicly available ChatGenerative Pre-trained Transformer (ChatGPT) in November 2022[1]. This milestone was quickly followed by the unveiling of similar models like Google's Bard[2], Anthropic's Claude [3], alongside open-source variants such as Meta's LLaMA[4]. LLMs are a subset of foundation models [5] (see **Glossary**), that are trained on massive text data, can have billions of parameters[6], and are primarily interacted with via text. Fundamentally, a language model serves as a channel that receives text queries and generates text in return[7]. LLMs can be adapted to a wide range of language-related tasks beyond their primary training objective. Their popularity has led to increasing interest in the medical field, and they have been applied to various tasks like note-taking[8], answering medical exam questions [9,10], answering patient questions[11], and generating clinical summaries[8]. Despite their versatility, LLM's behaviors are poorly understood[7], and they have the potential to produce medically inaccurate outputs[12] and amplify existing biases [13,14].

Evidence suggests that interest in LLMs is growing among physicians [8,15,16], and institutional partnerships are on the rise. Examples include its use in a training module for medical residents at Beth Israel Deaconess Medical Centre (Boston, USA)[17] and a partnership with EPIC, a major electronic health records provider, to integrate GPT-4 into their services[18]. As these models gain traction, it is essential for physicians to understand what LLMs are, their development, existing models, their current and potential applications, and the associated pitfalls when utilized in medicine. In this review, we will give an overview of how LLMs are trained as this background is instrumental in exploring their applications and drawbacks, describe previous ways that LLMs have been applied in medicine, and discuss both the limitations and potential for LLMs in medicine. Additionally, we provide several tutorial-like use cases to allow healthcare practitioners to trial some of the capabilities of one such model, ChatGPT using GPT-3.5.

**Table of Glossary**:

| Term | Definition |
| --- | --- |
| Neural networks (NN) | Systems inspired by the neuronal connections in the brain, that are capable of learning, recognizing patterns, and making predictions on tasks without explicit programming. They are the building blocks of many modern machine learning (deep learning) algorithms. |
| Foundation Model (FM) | A large-scale NN model trained on vast data to develop broad learning capabilities which can be fine-tuned for specific tasks. An FM can be fine-tuned to generate reports, answer medical questions. |
| Generative AI | Models trained on large datasets and can produce seemingly novel realistic content. This can be audio, visual, or text. |
| Large Language Models (LLMs) | Artificial intelligence models trained on an enormous amount of text data. LLMs are capable of generating human-like text and learning relationships between words. |
| Transformer Architecture | A deep learning model architecture that relies on self-attention mechanisms, by differentially weighting the importance of each part of the model's input. This makes it particularly useful for language tasks. |
| Attention | This is a mechanism within the Transformer architecture that enables the differential weighing mentioned above. |
| Parameters | These are values that are learned during the training process of a model. |
| Self-supervised learning | A form of training a model where it learns from unlabeled data, but utilizes the input data as its own supervision. A popular example is predicting the next word in a sentence. |
| Tokenization | This is a pre-training process in which text is converted into smaller units which can be like a character or a word, before being fed into the model. For example, hypertension can be tokenized into the following 'hy', 'per', 'tension'. |
| Pre-training | This is the initial phase of training a model on a large dataset before fine-tuning it on a task-specific dataset. The parameters are updated in the training process. |
| Fine-tuning | This refers to further training a pre-trained model on a specific task and adjusting the pre-existing parameters to achieve better performance for a particular task. |
| Zero-shot prompting | Using language prompts to get a model to perform specific tasks without having seen explicit examples of those tasks. |
| Few-shot prompting | In this case, the model is provided with some examples of the task, hence the name 'few'. |
| Instruction-tuning | Here, a pre-trained model is refined by providing explanations (instructions) on how to perform a task, alongside labeled examples that demonstrate the training objective/desired behavior. |
| Multi-modal LLMs | LLMs capable of processing and generating different types of data such as text, images, and audio. They are an emerging form of LLM with a wide range of applications in medicine. |
| In-Context Learning | This is the ability of a model to understand and generate appropriate responses based on the context of a given input. |
| Bias (in AI) | Systematic errors in the output of a model due to flawed assumptions in the machine learning process. This is usually from the data the model is trained on and can also be accentuated in the fine-tuning process. |
| MedMCQA | This is an open-source dataset of question-answer (QA) pairs that contains a collection of high-quality multiple-choice questions (over 194,000) from the AIIMS & NEET PG medical entrance examinations. It is one of the datasets used to develop and evaluate medical-related LLMs. |
| PubMedQA | This is like the MedMCQA, but focused on biomedical research QA pairs. They are also useful for developing and evaluating LLMs. |

## 2. Architecture of LLMs

LLMs rely on the 'Transformer' architecture [19,20]. This architecture leverages an 'attention' mechanism which uses multi-layered neural networks to help LLMs comprehend context and learn meaning within sentences and long paragraphs[6]. Akin to how a physician identifies important details of a patient's case while ignoring extraneous information, this mechanism enables LLMs to 'learn' important relationships between words while ignoring irrelevant information.

The training of these complex models involves billions of parameters, and has been made possible by recent advancements in computational power and model architecture[5] *(Figure 1)*. For example, GPT-3 was trained on vast data sources and reportedly has about 175 billion parameters[21], while the open source LLaMA family of models have 7 to 70 billion parameters [4,22]. The first step of LLM training, known as pre-training, is a self-supervised approach that involves training on a large corpus of unlabeled data, such as internet text, Wikipedia, Github code, social media posts, and BooksCorpus[4–6]. Some are also trained on proprietary datasets containing specialized texts like scientific articles[4]. The training objective is usually to predict the next word in a sentence, and this process is resource-intensive[23]. It requires the conversion of the text into tokens before they are fed into the model[24]. The result of this step is a base model that is in itself simply a general language-generating model, but lacks the capacity for nuanced tasks.

This base model then undergoes a second phase, known as fine-tuning[24]. Here, the model can be further trained on narrower datasets like medical transcripts for a healthcare application, or legal briefs for a legal assistant bot. This fine-tuning process can be augmented with a Constitutional AI approach[25], which involves embedding predefined rules or principles directly into the model's architecture. Also, this phase can be enhanced with reward training [6,26], where humans score the quality of multiple model outputs, and a reinforcement learning from human feedback (RLHF) approach[26], which employs a comparison-based system to optimize the model responses. This step, which is less computationally expensive, albeit human-intensive, adjusts the model to perform a specific task with controlled outputs. The fine-tuned model from this phase is what is deployed in flexible applications like a chatbot.

LLMs' adaptability to unfamiliar tasks[27] and apparent reasoning abilities[28] are captivating. However, unlocking their full potential in specialized fields like medicine requires even more specific training strategies. These strategies could include direct prompting techniques like few-shot learning [21,29], where a few examples of a task at test time guide the model's responses, and zero-shot learning [30–32], where no prior specific examples are given. More nuanced approaches such as chain-of-thought prompting[33], which encourages the model to detail its reasoning process step by step, and self-consistency prompting[34], where the model is challenged to verify the consistency of its responses, also play important roles.

Another promising technique is instruction prompt tuning, introduced by Lester et al.[35], which provides a cost-effective solution to update the model's parameters, thereby improving performance in many downstream medical tasks. This approach offers significant benefits over the few-shot prompt approaches, particularly for clinical applications as demonstrated by Singhal et al[12]. Overall, these methods augment the core training processes of fine-tuned models and can enhance their alignment with medical tasks as recently shown in the Flan-PaLM model[12]. As these models continue to evolve, understanding their training methodologies will serve as a good foundation for discussing their current capabilities and future applications.

*Figure 1. Overview of LLM training process. LLMs 'learn' from more focused inputs at each stage of the training process. The first phase of this learning is pre-training, where the LLM can be trained on a mix of unlabeled data and proprietary data without any human supervision. The second phase is fine-tuning, where narrower datasets and human feedback are introduced as inputs to the base model. The fine-tuned model can then enter an additional phase, where humans with specialized knowledge implement prompting techniques that can transform the LLM into a model that is augmented to perform specialized tasks.*

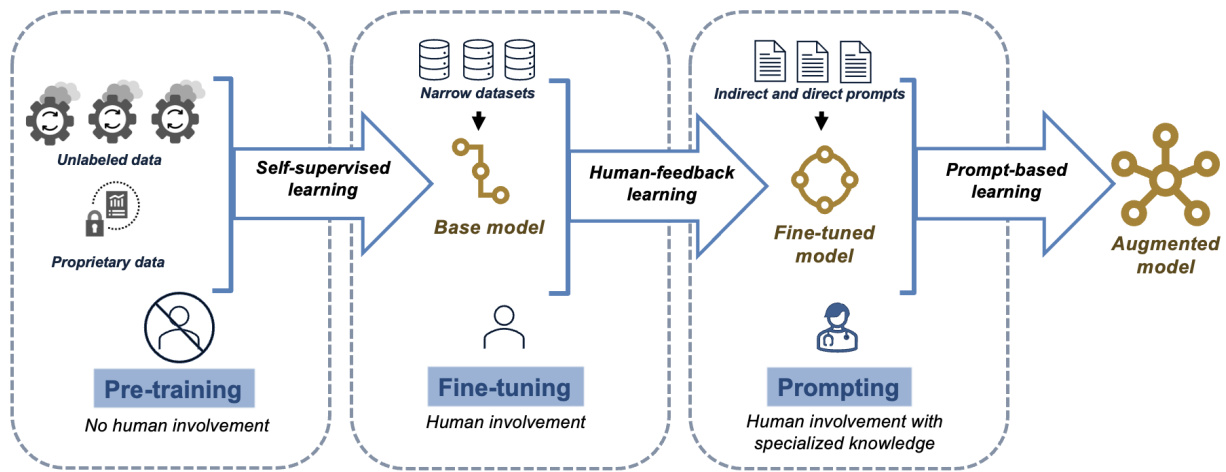

## 3. Overview of current medical-LLMs

Prior to the emergence of LLMs, natural language processing challenges were tackled by more rudimentary language models like statistical language models (SLM) and neural language models (NLM)[6], which had significantly fewer parameters and trained on relatively small datasets. These predecessor models lacked the emergent capabilities of modern LLMs, such as reasoning and in-context learning. The advent of the Transformer architecture was a pivotal point, heralding the age of multifaceted LLMs we see today. Often, specialized datasets are used to evaluate an LLMs' performance on medical tasks, typically deploying an array of QA tools like MedMCQA, PubMedQA[12,36] (in Glossary), and more novel ones like MultiMedBench[37].

In this section, we will provide an overview of general-purpose LLMs, with a specific emphasis on those that have been applied to tasks within the medical field. Additionally, we'll delve into domain-specific LLMs, referring to models that have been either pre-trained or uniquely fine-tuned using medical literature *(Figure 2)*.

- **Generative Pre-trained Transformers (GPT)**: Arguably the most popular of the general Large Language Models are those that belong to the GPT lineage, largely due to their chat-facing product. Developed by OpenAI in 2018[38], the GPT series has significantly scaled in recent years, with the latest version, GPT-4, speculated to possess significantly more parameters than its predecessors. The evolution in parameters from a mere 0.12B in GPT-1 speaks to the enormous strides made in this area. GPT-4 represents a leap forward in terms of its ability to handle multi-modal input such as images, text, and audio - an attribute that aligns seamlessly with the multi-faceted nature of medical practice. Novel prompting techniques were introduced with the GPT models, paving the way for the popular ChatGPT product, which is based on GPT-3.5 and GPT-4. ChatGPT has demonstrated its utility in various medical scenarios discussed later in this paper[8,39–41]. Certain studies have concentrated on evaluating the healthcare utility of ChatGPT and InstructGPT, while others have focused on its fine-tuning for specific medical tasks. For instance, Luo et al. introduced BioGPT, a model that utilized the GPT-2 framework pre-trained on 15 million PubMed abstracts for tasks including question answering (QA), relation extraction, and document classification[39]. Their model outperformed the state-of-the-art models across all evaluated tasks. In a similar vein, BioMedLM 2.7B (formerly known as PubMedGPT), pre-trained on both PubMed abstracts and full texts[42], demonstrates the continued advancements in this field. Some researchers have even leveraged GPT-4 to create multi-modal medical LLMs, reporting promising results[43,44].
- **Bidirectional Encoder Representations from Transformers (BERT):** Another prominent category of language models that warrant discussion stems from the BERT family. First introduced by Devlin and colleagues, BERT was unique due its focus on understanding sentences through bidirectional training of the model, compared to previous models that used context from one side[45]. For medical tasks, researchers have developed domain-specific versions of BERT tailored to scientific and clinical text. BioBERT incorporates biomedical corpus data from PubMed abstracts and PubMed Central articles during pre-training[46]. PubMedBERT follows a similar methodology using just PubMed abstracts[47]. ClinicalBERT adapts BERT for clinical notes, trained on the large MIMIC-III dataset of electronic health records[48]. More recent work has focused on enhancing BERT for specific applications. BioLinkBERT adds entity linking to connect biomedical concepts in text to ontologies[49]. These extensions showcase how baseline BERT architectures can be customized for medicine. With proper tuning BERT-based LLMs have demonstrated potential to augment various medical tasks.

- **Pathways Language Model (PaLM)**: Developed by Google, this model represents one of the largest LLMs to date. Researchers first fine-tuned PaLM for medical QA, creating the Flan-PaLM[50] which achieved state-of-the-art results on QA benchmarks. Building on this, the Med-PaLM model was produced via instruction tuning, demonstrating strong capabilities in clinical knowledge, scientific consensus, and medical reasoning[12]. This has recently been extended to create a multimodal medical LLM[37]. PaLM-based models underscore the utility of large foundation models fine-tuned for medicine.
- Beyond big tech companies, other proprietary and open source medical LLMs have emerged. Models trained from scratch on clinical corpora, such as GatorTron[51], have shown improved performance on certain tasks compared to general domain LLMs. Claude, developed by Anthropic, has been evaluated on medical biases and other safety issues for clinical applications[13]. Active open-source projects are also contributing to the medical LLMs field. For example, PMC-LLaMA leverages the LLaMA model and incorporates biomedical papers in its pre-training [52]. Other popular base models like DRAGON[53], Megatron[54], and Vicuna[55] have enabled development of multimodal LLMs incorporating visual data[43].

Overall, domain-specific pre-training on medical corpora produces models that excel on biomedical tasks compared to generalist LLMs (with some exceptions like GPT-4). However, fine-tuning approaches for adapting general models like BERT and GPT-3 have achieved strong results on medical tasks in a more computationally efficient manner. This is promising given the challenges of limited medical data for training.

As LLMs continue to scale up with larger parameter counts, they appear likely to implicitly learn useful biological and clinical knowledge, evidenced by models like Med-PaLM demonstrating improved accuracy, calibration, and physician-like responses. While not equal to real clinical expertise, these characteristics highlight the growing potential of LLMs in medicine and healthcare.

*Figure 2. Current LLMs in medicine. Currently there are general purpose and biomedical LLMs used for medical tasks. While GPT by OpenAI, BERT by Devlin and colleagues, and PaLM by Google have led the development of LLMs with applications in medicine, other proprietary and open source LLMs also exist in this space. Circle sizes reflect the model size and the number of parameters used to build the models. LLMs with applications in medicine vary widely in how they were trained. BioMedLM 2.7B by GPT was trained on the corpus of PubMed articles and abstracts, for example, whereas ClinicalBERT was trained specifically on electronic health records. These differences in training and development can have important implications for how LLMs perform in certain medical scenarios.*

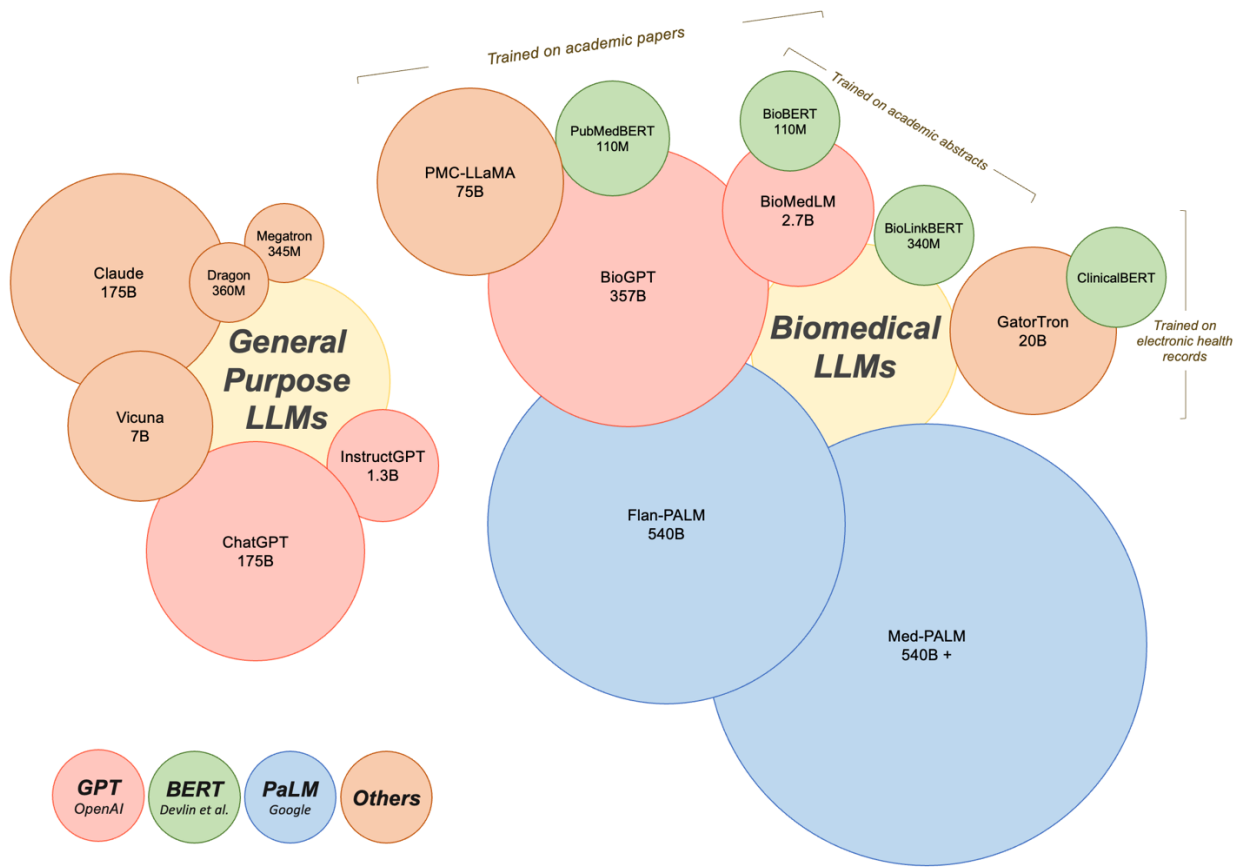

# 4. LLM in medical tasks:

*4.1 Overview of LLM in medicine*
Given the rapid advances in LLMs, there has already been an incredible amount of research conducted exploring the usage of LLMs in medicine, ranging from answering patient questions in cardiology[10] to serving as a support tool in tumor boards[15], to aiding researchers in academia [56,57]. A brief search in PubMed for "ChatGPT" revealed 800-plus results, showing the rapid exploration and adoption of this technology.

Prior to ChatGPT, many patients have been using the internet to learn more about their health conditions [58,59]. As ChatGPT surfaced, one of the first-line uses for the language model was answering patient questions. In cardiology, researchers have found that ChatGPT was able to adequately respond to prevention questions, suggesting that LLMs could help augment patient education and patient-clinician communication[10]. Similarly, researchers have explored ChatGPT's responses to common patient questions in hip replacements[11], radiology report findings[60], and management of venomous snake bites[61]. Additionally, there has been interest in using ChatGPT to aid with translating medical texts and clinical encounters, with an objective of improving patient communication and satisfaction[62]. These findings suggest a potential for bridging gaps in patient education; however, additional testing is needed to ensure fairness and accuracy.

In addition to augmenting patient education, researchers are exploring LLM's use as clinical workflow support tools. One study evaluated ChatGPT's recommendations for next step management in breast cancer tumor boards, which are frequently composed of the most complex clinical cases[15]. Other studies explored the use of ChatGPT in responding to patient portal messages[40], creating discharge summaries[41], writing operative notes[63], and generating structured templates for radiology [60]. While these studies suggest opportunities for mitigating the documentation burden facing physicians, rigorous real world evaluation should be completed prior to any clinical use.

Aside from uses in clinical medicine, LLMs are being utilized in medical education and academia. Multiple researchers have explored LLM's ability to conduct radiation oncology physics calculations[64], answer medical board questions in USMLE style[9], and respond to clinical vignettes[65]. The ability of this technology to adequately achieve passing scores on these medical exams raises questions on the need to revise medical curriculum and practices [66]. Other programs have started exploring using LLM's generative ability to create multiple choice questions for student exams[67]. In academia, there is a rise in exploring LLM's ability to aid researchers, ranging from topic brainstorming[57] to writing journal articles [56,68], resulting in a rising debate on the ethics and usage of LLMs in academic writing.

*4.2 Proposed tasks of LLM in medicine:*
As evident by the abundance of work already done with LLMs, there is an ubiquitous amount of tasks that this technology can aid clinicians with, ranging from administrative tasks to gathering and enhancing medical knowledge (**Table 1**).

**Table 1**. Analysis of possible large language models tasks in medicine

| Task | Potential Pitfalls | Mitigation Strategies |
|---|---|---|
| Administrative:<br>- Write insurance authorization letters<br>- Summarize medical notes<br>- Aid medical record documentation<br>- Create patient communication (email/letter/text) | Lack of HIPAA compliance: No publicly available model is currently HIPAA compliant, and thus PHI cannot be shared with the models. | Integrate LLMs within electronic health record systems. |
| Augmenting knowledge:<br>- Answer diagnostic questions<br>- Answer questions about medical management<br>- Create and translate patient education material | Inherent bias: Pre-trained data models used for diagnostic analyses will introduce inherent bias. | Create domain-specific models that are trained on carefully curated datasets. Always include a human in the loop. |
| Medical education:<br>- Write recommendation letters<br>- Create new exam questions and case-based scenarios<br>- Generate summaries of medical text at a student level | Lack of personalization: LLMs are generated from prior work already published, resulting in repetitive and unoriginal work. | Educate clinicians and users in using LLM tools to augment their work rather than replace.<br><br>Encourage understanding how the technology works to mitigate unrealistic expectations of output. |
| Medical research:<br>- Generate research ideas and novel directions<br>- Write academic papers<br>- Write grants | Ethics: There has been an incredible amount of discussion among the scientific community on the ethics of using ChatGPT to generate scientific publications. This also raises the question of accessibility and the potential difficulties of future access to this technology. | Engage in conversation to increase accessibility of this technology to prevent widening gaps in research disparities. |

## 5. Limitations and mitigation strategies:

While researchers have demonstrated the feasibility of LLM's use in medicine, there are also many limitations to these preliminary studies, emphasizing the need for future research and analysis. As discussed briefly above in *Table 1*, there are many potential pitfalls that clinicians using this technology need to be aware of. Key challenges posed by LLM include issues related to accuracy, bias, model inputs/outputs and privacy/ethical concerns. By understanding and addressing these limitations, researchers can foster responsible development and usage of these models to create a more equitable and trustworthy ecosystem.

*5.1 Accuracy issues and dataset bias*
Models are only as accurate as the datasets that are used to train them, resulting in a reliance on the accuracy and completeness of the data. LLMs are trained on large datasets that have long surpassed the ability of human teams to manually quality check. This results in a model that is trained on a nebulous dataset that may further decrease user trust in these algorithms. Due to the inability to quality check the dataset, there is often overlap between the training and testing datasets, which results in overprediction of the accuracies of the models [24]. Authors of GPT-3, the base model that ChatGPT was built on, reported that after training the model, they discovered some overlap in their dataset, but could not afford to retrain the model [21]. In addition, frequently factual information that was used to train the models can become outdated, and it is nontrivial to retrain the model on updated information.

It is also important to note that ChatGPT and many of the other LLMs are not trained on specifically curated medical datasets, but rather on a broad range of inputs, ranging from news articles to literary works, that allow models to capture linguistic patterns and features. This results in poor performance in domain-specific questions, including medical applications[69].

Aside from the source of the datasets, it is important to explore the specific characteristics of the datasets. Models frequently enhance and reinforce structural biases that are found in the training datasets. Multiple groups have exhibited that models are promoting race-based medicine practices that have long been scientifically refuted. When answering questions about eGFR calculations, multiple LLMs tried to justify race-based medicine with false assertions about Black people having different muscle mass and thus higher creatinine levels[13]. Others have found that LLMs associate phrases referencing people with disabilities with more negative sentiment words, and that gun violence, homelessness, and drug addiction are overrepresented in texts discussing mental illness[70]. In another scenario, multiple LLMs were asked to provide analgesia choices for chest pain for white patients and black patients, resulting in weaker analgesic recommendations for black patients[71].

A way to mitigate the accuracy and bias of these models is to train these models on domain-specific datasets. Work has been done to fine-tune LLMs in radiology, creating a system that leverages radiology reports during model creation, ultimately increasing the performance of the models for radiology-specific tasks [72]. Aside from fine-tuning previously trained LLMs, there has been work done to create models from scratch using electronic health record data, coined as clinical foundation models [73]. These models were shown to have better predictive performance, require less labeled data, more effectively handle multimodal data, and offer novel interfaces for human-AI interaction [73]. Clinicians can also aid developers in decreasing dataset bias by working to gather more diverse datasets for these models to train on, by conducting outreach to underrepresented patient populations.

*5.2 Weak input, poor output, and change over time*
The inputs of LLMs are very fickle; very small changes in the input wording results in dramatic changes in the output. Frequently, these variations in prompt syntax often occur in ways that are unintuitive to the users [24]. This causes difficulties with ensuring consistency when using LLMs in a healthcare setting.

In addition to the fickle nature of inputs, models frequently generate "hallucinations", where the model produces nonsensical and factually incorrect responses [74,75]. This is exacerbated when insufficient information is provided in the prompt, a scenario that is frequently seen in healthcare. Researchers prompted LLMs to summarize documents, and found substantial amounts of hallucinations in the summaries, where the model will insert grossly inaccurate information not found in the original document inputs [76]. In addition, these language models frequently utilize confident language in the output which could lead users to blindly trust LLMs outputs despite incorrect information[77].

Finally, because many LLMs take inputs as truthful, it attempts to generate an output that fits the user's assumption, rather than offering factual corrections or clarifying prompts[71]. This inherently raises challenges for use cases in medicine, where researchers and clinicians may exacerbate misconceptions, worsening confirmation biases. To help mitigate some of these limitations, clinicians and researchers should be well-versed in prompt engineering to encourage accurate and sensible use of this new technology [78].

Aside from the fickle nature of inputs and unpredictable nature of the outputs, LLMs are also evolving rapidly and unpredictably over time [79]. This makes it challenging to incorporate these models into larger workflows. Given the novelty of this technology, much work still remains to observe long-term trends and analyses. Utilization of this technology in healthcare should not be without careful oversight.

*5.3 Privacy and ethical concerns:*
Frequently, personally identifiable information (PII) have been found within pre-training datasets in earlier LLMs, including phone numbers and email addresses [80]. Even if the datasets are completely devoid of PII, privacy violations can occur due to inference. LLMs may make correct assumptions about a patient based on correlational data about other people without access to private data about that particular individual. In other words, LLMs may attempt to predict the patient's gender, race, income, or religion based on user input, ultimately violating the individual's privacy [77]. Researchers have already shown that patient-entered text, in the form of Twitter accounts, are able to accurately predict alcoholism recovery[81]. LLMs are text-based models that may have these abilities as well.

Aside from the above privacy issues, there are ethical concerns with LLM use in medicine. Even when assuming non-malicious users of these models, there are unfortunately opportunities for the models to generate harmful content. For example, when disclosing difficult diagnoses in medicine, there are steps in place that can help patients cope and provide support. With the rise of LLMs in medicine, patients may inadvertently be exposed to difficult topics that can cause severe emotional harm. While this problem is not unique to language models, as patients have other means to access information (E.g. Google), LLMs produce a greater risk given the conversation-like structure of these publicly-available models. Many of these models are human-enough, but frequently lack the ability for additional personalized emotional support.

Another ethical concern is the rising difficulty in distinguishing between LLM-generated and human-written text, which could result in spread of misinformation, plagiarism, and impersonation. For example, while the use of LLMs to aid clinicians in administrative tasks can help decrease document burden, this could also result in malicious use by others to generate false documents.

To help mitigate these privacy and ethical concerns, regular auditing and evaluation of LLMs can help identify and address these ethical concerns. There have been calls recently in the AI research community to develop regulations for LLM use in medicine[82]. In addition, care must be taken in the selection of training datasets, especially when using medicine domain-specific datasets to ensure adequate handling of sensitive data.

## 6. Tutorial with ChatGPT

In this section, we will walk through some of the use cases that highlight both the benefits and pitfalls of one LLM, OpenAI's ChatGPT *(Table 1)*. To complete this tutorial section, individuals will need to visit https://openai.com/ and create a free account for ChatGPT.

*6.1 Insurance authorization letters*
We will start with creating authorization letters, which is quite common in the clinical setting. The prompt for this exercise is: "Can you write an authorization letter for Aetna insurance for a total left hip replacement procedure in a patient with osteoarthritis of the left hip? Please ensure to be accurate as this service is not typically covered by insurance." It is important to not include personally identifiable information and protected health information when utilizing this tool. From the model output in Figure 3a, you can see that sections that require identifiable details are filled with placeholders. You can tweak the prompts or add more details as required and see how that changes the outputs.

*6.2 Exploring hallucinations via basic patient handouts*
To create a basic patient handout, we use the prompt, "Create a one-page patient handout about acid reflux to be used in a physician's office. Use accessible language that is at a 5th grade level." When we use this prompt, we get the output seen in Figure 3b. The information in this generated handout appears to be accurate. However, as mentioned before, these models can "hallucinate"; in this case, the "sources" listed on this handout do not exist. Now repeat the same prompt and look at the output, do you get the same result? As mentioned previously, these models are stochastic and do not give the exact same output (in some cases, you can change a "temperature" parameter which controls stochasticity).

*6.3 Exam questions and case-based scenarios*
Here, we used GPT-4, the more advanced model (which is available through a paid subscription). For this scenario, we use the prompt, "Create a USMLE Step 3 question that tests the concept of digoxin side effects." The output from this prompt is shown in Figure 3c. The output is clear and you can see the model attempts to provide an explanation for the answer. Although, in reality the NBME Step 3 questions could be more vague. Try this with different prompts on various medical concepts and evaluate what you get. Feel free to use the free ChatGPT version. You are likely to see some hallucinations here when the model tries to explain the answer. You can also ask it to use a different question style and see if it meets your expectations.

*6.4 Honing model outputs via academic paper outlines*
To use this technology to help with medical research, we use the prompt, "Create an outline for a review paper on the implications of social media in healthcare." While the output (Figure 3d) creates a comprehensive outline for researchers to follow, we can see that the first iteration of the outline is quite generic. You can then continue to "converse" with the model to provide more details and enhancements to improve the outline and the outputs. We use the prompt "Include a literature review of prior works in the introduction", which allows the model to hone in on creating language to aid researchers in writing their papers. Explore asking ChatGPT to provide more or less detail to create the outline for your paper.

*Figure 3. ChatGPT's output to the following prompts.* a) Can you write an authorization letter for Aetna insurance for a total left hip replacement procedure in a patient with osteoarthritis of the left hip? Please ensure to be accurate as this service is not typically covered by insurance b) Create a one-page patient handout about acid reflux to be used in a physician's office. Use accessible language that is at a 5th grade level. c) Create a USMLE Step 3 question that tests the concept of digoxin side effects. d) Create an outline for a review paper on the implications of social media in healthcare

c) USMLE Step 3 Question:

A 76-year-old female with a history of heart failure with reduced ejection fraction presents to the clinic with complaints of fatigue, loss of appetite, blurred vision, and intermittent confusion for the past week. She denies chest pain, shortness of breath, or cough. Her medications include carvedilol, lisinopril, furosemide, and digoxin, which was started 2 months ago. On examination, her blood pressure is 125/75 mmHg, heart rate 52/min, and respiratory rate 16/min. An electrocardiogram (ECG) shows atrial fibrillation with a ventricular response of 52/min and frequent premature ventricular contractions (PVCs). Laboratory findings show a digoxin level of 2.8 ng/mL.

Which of the following is the most appropriate next step in management?

A) Discontinue digoxin and administer digoxin-specific antibody fragments.
B) Increase the dose of furosemide.
C) Start amiodarone for rhythm control.
D) Increase the dose of carvedilol.
E) Discontinue lisinopril.

Answer: A) Discontinue digoxin and administer digoxin-specific antibody fragments.

Explanation:
The patient's presentation is consistent with digoxin toxicity. Digoxin has a narrow therapeutic window and side effects are relatively common, particularly in elderly patients. Symptoms of digoxin toxicity are variable and include gastrointestinal (anorexia, nausea, vomiting, diarrhea), neurological (fatigue, weakness, confusion, visual disturbances like yellow vision) and cardiac manifestations (bradycardia, atrioventricular block, atrial tachyarrhythmias, and PVCs). This patient's decreased appetite, blurred vision, confusion, bradycardia, and PVCs on ECG are consistent with digoxin toxicity. Her digoxin level further confirms the diagnosis. The most appropriate management of acute digoxin toxicity with life-threatening features (e.g., significant bradycardia, ventricular arrhythmias) is discontinuation of digoxin and administration of digoxin-specific antibody fragments.

d) Title: Implications of Social Media in Healthcare: A Comprehensive Review

I. Introduction
A. Brief overview of the rise of social media in the healthcare sector
B. Importance of studying the implications of social media in healthcare
C. Purpose and scope of the review paper

II. Evolution of Social Media in Healthcare
A. Historical context: emergence of social media platforms
B. Adoption and integration of social media in healthcare organizations
C. Examples of popular social media platforms used in healthcare (e.g., Twitter, Facebook, LinkedIn, Instagram)
D. Trends in patient engagement and empowerment through social media

III. Benefits of Social Media in Healthcare
A. Patient Education and Awareness
1. Dissemination of accurate medical information
2. Health campaigns and awareness initiatives

B. Healthcare Communication and Collaboration
1. Facilitating communication among healthcare professionals
2. Cross-disciplinary collaboration and knowledge sharing

C. Patient-Provider Interaction
1. Telemedicine and virtual consultations
2. Enhancing patient engagement and satisfaction

D. Data Mining and Public Health Surveillance
1. Analyzing public health trends through social media data
2. Early detection of disease outbreaks and health concerns

E. Support Communities and Patient Empowerment
1. Online support groups for patients with chronic conditions
2. Empowering patients to manage their health

IV. Challenges and Concerns
A. Patient Privacy and Data Security
1. Risks of sharing personal health information on social media
2. Ensuring HIPAA compliance and safeguarding patient data

B. Misinformation and Trustworthiness
1. Spread of false medical information
2. Strategies for combating misinformation

## 7. Future of LLM in medicine:

LLMs are currently at the forefront of AI innovation in medicine, with a surge of new developments being introduced regularly. Their potential to improve care delivery and alter the practice of medicine is notable. Here we discuss future developments for LLMs in medicine could look like, drawing on both currently emerging trends and future conjectures.

**Technological Advancements**: The integration of multiple data types into LLMs, referred to as multi-modality, is an emerging trend[83] with significant implications for healthcare[84]. Initially introduced by GPT-4, this property has been further developed for medicine with a proof-of-concept generalist medical AI called Med-PaLM Multimodal (Med-PaLM M)[37]. Recent studies such as LLaVa-Med[85], SkinGPT4[44], and MiniGPT4[43] provide compelling evidence for the effectiveness of multi-modal LLMs, which are poised to gain prevalence in healthcare due to the multi-faceted nature of medical data that spans text, images, audio, and genetics. Also, more advanced models with better architectures and longer context length (which enables models to maintain coherence) could lead to more accurate responses for medical tasks.

Simultaneously, progress in minimizing resource requirements for LLMs is likely to democratize access and benefit physicians in resource-limited settings, enabling them to train LLMs for their own clinical and research tasks. By extension, this could reduce racial and gender bias[14] in model outputs as more robust models are developed. Moreover, the reduction in demand for compute resources could pave the way for institution-specific LLMs - models trained on data specific to a health institution, thereby reflecting its standard procedures, guidelines, and unique challenges. Such models have the potential to enhance productivity, reduce burnout, and improve patient care.

**Accessibility and Equity**: The creation of synthetic medical data by leveraging the generative capabilities of LLMs also offers a promising approach to address the challenges associated with the scarcity of medical research data. For example, more diverse medical data could be available for training AI models. This can lead to more inclusive and equitable medical research.

**Regulatory Considerations**: From a regulatory standpoint, it is imperative to establish standard frameworks for validating LLMs across clinical tasks, while ensuring fairness. This is particularly crucial in medicine, where inaccurate model outputs can have severe consequences and lead to patient harm. LLM governance structures need to evolve to protect patient privacy, and address issues like model transparency, fairness, and accuracy[82].

**Future Research Directions**: Research into LLM explainability, where the model provides logical reasoning for its decision-making or question-answering process, is essential in medicine and should be prioritized. Models that can elucidate their reasoning are more likely to gain acceptance and trust from physicians. Also, standardized holistic metrics for the assessment of LLM abilities in medicine are necessary to improve widespread adoption. These metrics should be holistic and cover clinical accuracy, reasoning, bias, and fairness.

In summary, the future of LLMs in medicine will likely feature advancements that enhance their utility as supportive tools for healthcare workers, not replacement. These developments could play a crucial role in addressing challenges related to healthcare shortages and inefficiencies.

## 8. Conclusion

LLMs have risen in popularity as models become more widely available for public use. With technological advancements and spread in popularity, potential opportunities exist for application of LLMs in the medical field. Multiple tech companies have already developed models trained with the intention to perform medical tasks. There are several areas of medicine where LLMs could be employed, such as completing administrative tasks (e.g.: summarizing medical notes), augmenting clinician knowledge (e.g.: translating patient materials), medical education (e.g.: creating new exam questions), and medical research (e.g.: generating novel research ideas). Despite these opportunities, many notable challenges with LLMs remain unresolved, limiting the implementation of these models in medicine. Issues surrounding underlying biases in datasets, data quality and unpredictability of outputs, and patient privacy and ethical concerns make innovations in LLMs difficult to translate to adoption in healthcare settings in their current form. Physicians and other healthcare professionals must weigh potential opportunities with these existing limitations as they seek to incorporate LLMs into their practice of medicine.